\documentclass[letterpaper, 10 pt, conference]{ieeeconf}  
\usepackage[utf8]{inputenc}
\usepackage[english]{babel}
\usepackage[T1]{fontenc}
\usepackage{hyperref}
\usepackage{url}
\usepackage{booktabs}
\usepackage{amsfonts}
\usepackage{amsmath}
\usepackage{mathtools}
\usepackage{nicefrac}
\usepackage{microtype}
\usepackage{graphicx}
\usepackage{tikz}
\usepackage{subcaption}
\usepackage{nicefrac}
\usepackage{pgfplots}
\usepackage{multirow}
\usepackage{float}
\usepackage[ruled,lined,linesnumbered,english]{algorithm2e}
\SetKwFor{ForEach}{for each}{do}{end}
\SetKwFor{ForAll}{for all}{do}{end}

\usetikzlibrary{intersections}
\usetikzlibrary{patterns}
\usetikzlibrary{calc}
\usetikzlibrary{decorations.pathreplacing}
\usetikzlibrary{fadings}
\usetikzlibrary{positioning,fit,calc}
\usetikzlibrary{arrows,shadows}
\usetikzlibrary{shapes.geometric}

\definecolor{matchnet}{RGB}{250,192,144}
\definecolor{scorenet}{RGB}{195,214,155}

\tikzset{
  network/.style={
    rectangle, draw,
    text width=6em, text centered,
    minimum height=4em,
    fill=white
  },
  frame/.style={
    rectangle, draw,
    text width=6em, text centered,
    minimum height=4em,
    fill=white
  },
  line/.style={
    draw, -latex',rounded corners=3mm,
  },
  textnode/.style={
    rectangle,
    text width=6em, text centered,
    minimum height=4em,inner sep=0pt,
  },
}
\tikzstyle{input}=[draw,rounded corners=10pt,fill=green!50, text width=8.5em, text centered, minimum height=4em]
\tikzstyle{mlp}=[draw,rounded corners=10pt,fill=blue!20, text width=8.5em, text centered, minimum height=4em]
\tikzstyle{conv}=[draw,rounded corners=10pt,fill=yellow!30, text width=8.5em, text centered, minimum height=4em]
\tikzstyle{max}=[draw,dashed,rounded corners=10pt,fill=red!15, text width=8.5em, text centered, minimum height=4em]
\tikzstyle{skip}=[draw,dashed,rounded corners=10pt,fill=brown!15, text width=8.5em, text centered, minimum height=4em]
\tikzstyle{output}=[draw,rounded corners=10pt,fill=green!50, text width=8.5em, text centered, minimum height=4em]

\tikzstyle{stateTransition}=[->, very thick]

\newcommand{\abs}[1]{\left\lvert#1\right\rvert}

\newcommand{\binvarset}{\mathbf{y}}
\newcommand{\binvar}{y}
\newcommand{\trainset}{\mathcal{X}}
\newcommand{\obsvarset}{\mathbf{x}}
\newcommand{\obsvar}{x}
\newcommand{\weightset}{\mathbf{W}}

\newcommand{\modelfun}{\theta}
\newcommand{\bx}{\mathbf{x}}
\newcommand{\by}{\mathbf{y}}

\newcommand{\raquel}[1]{}
\newcommand{\davi}[1]{}

\graphicspath{{imgs/}} 

\title{\LARGE \bf
End-to-end Learning of Multi-sensor 3D Tracking by Detection
}

\author{Davi Frossard \qquad Raquel Urtasun\\
Uber Advanced Technologies Group\\
University of Toronto\\
{\tt\small \{frossard, urtasun\}@uber.com}}

\begin{document}

\maketitle
\thispagestyle{empty}
\pagestyle{empty}

\begin{abstract}

  In this paper we propose a novel approach to  tracking by detection
that can exploit both cameras as well as LIDAR data to produce very accurate 3D trajectories.
Towards this goal, we formulate the  problem as a linear program that can be solved exactly, and learn convolutional networks for detection as well as matching in an end-to-end manner.  We evaluate our model in the challenging KITTI dataset and show very competitive results.

\end{abstract}

\section{Introduction}
\label{sec:introduction}

\begin{figure*}[t]
\centering
  \includegraphics[width=\textwidth]{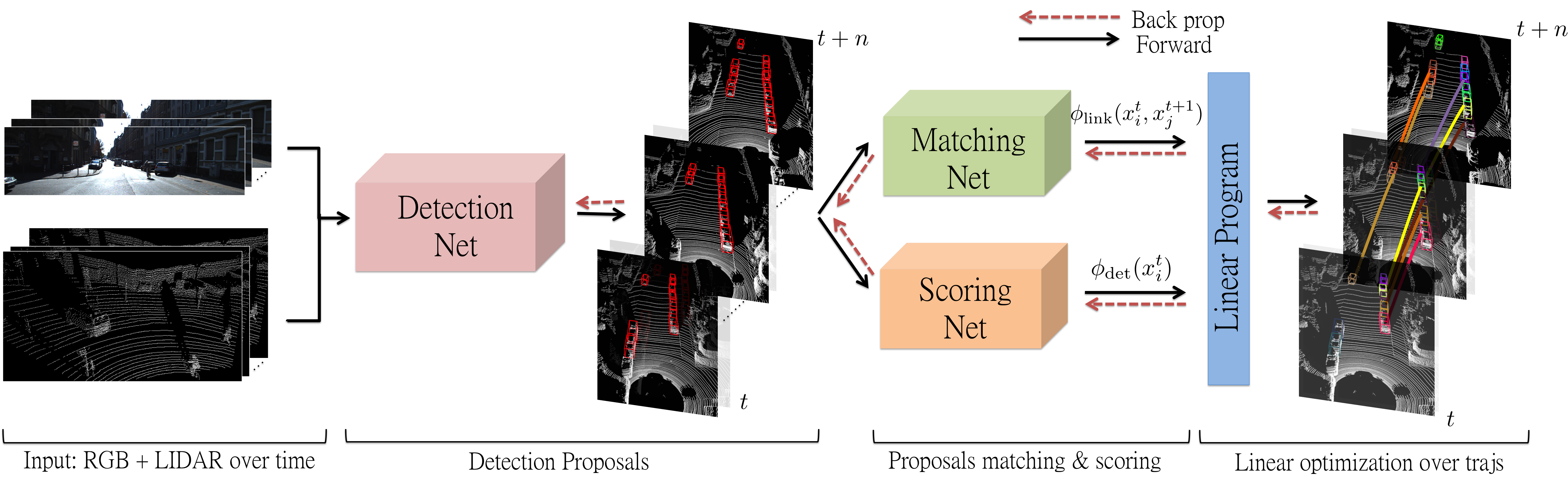}
  \caption{In this work, we formulate tracking as a system containing multiple neural networks that are interwoven together in a single architecture. Note that the system takes as external input a time series of RGB Frames (camera images) and LIDAR pointclouds. From these inputs, the system produces discrete trajectories of the targets. In particular, we propose an architecture that is end to end trainable while still maintaining explainability, we achieve this by formulating the system in a structured manner.}
  \label{fig:framework_viz}
\end{figure*}

One of the fundamental tasks in perception systems for autonomous driving is to be able to track traffic participants. This task, commonly referred to as {\it Multi-target tracking},  consists on  identifying how many objects there are in each frame, as well as link their trajectories over time. Despite many decades of research, tracking is still an open problem. Challenges include dealing with object truncation, high speed targets, lighting conditions, sensor motion and  complex interactions between targets, which leads to occlusion and path crossing.


Most modern computer vision approaches to multi-target tracking are based on \textit{tracking by detection} \cite{onlinemcf}, where first a set of possible objects are identified via   object detectors. These   detections are then further associated over time in a second step by solving a discrete problem.
Both tracking and detection are typically formulated in 2D, and a variety of cues based on appearance and motion are exploited.

In robotics, \textit{tracking by filtering} methods are more prevalent, where the input is filtered in search of moving objects and  their state is predicted over time \cite{thrun2005probabilistic}.
LIDAR based approaches are the most common option for 3D tracking, since this sensor provides an accurate spatial representation of the world allowing for precise positioning  of the objects of interest. However, matching is more difficult as  LIDAR does not capture appearance well when compared to the richness of images. 




In this paper, we propose an approach  that can take advantage of both  LIDAR and camera data. 
Towards this goal, we formulate the problem as inference in a deep structured model, where the potentials are computed using convolutional neural nets. Notably, our matching cost of associating two detections exploits both appearance and motion via a siamese network that processes images  and motion representations via convolutional layers. 
Inference in our model can be done exactly and efficiently by a set of feedforward passes followed by solving a linear program. Importantly, our model is formulated such that it can be trained end-to-end to solve both the detection and tracking problems. We refer the reader to \autoref{fig:framework_viz} for an overview our approach. 

%


\section{Related Work}
\label{sec:related_work}

Recent works in multiple object tracking are usually done in two fronts: Filtering based and batch based methods. Filtering based methods rely on the Markov assumption to estimate the posterior distribution of the trajectories.   Bayesian or Monte Carlo filtering methods such as Gaussian Processes \cite{urtasun20063d}, Particle Filters and Kalman Filters \cite{thrun2005probabilistic} are commonly employed. One advantage of filtering approaches is their efficiency, which allows for real-time applications. However, they suffer from the propagation of early errors, which are hard to mitigate. To tackle this shortcoming, batch methods utilize object hypotheses from a detector (\textit{tracking by detection}) over entire sequences to estimate trajectories, which allows for global optimization and usage of higher level cues. Estimating trajectories becomes a data association problem, i.e., deciding from the set of detections which should be linked to form correct trajectories. The association   can be estimated with  Markov Chain Monte Carlo (MCMC) \cite{Collins2014,choi_pami13}, linear programming \cite{berclaz,shitrit} or with a flow graph \cite{mincostflow}.

Online methods have also been proposed in order to tackle the performance issue with batch methods \cite{onlinemcf,onlinecrf}. Milan et al. \cite{milan2016online} use Recurrent Neural Networks (RNN) to encode the state-space and solve the association problem.

Our work also expands on previous research on pixel matching, which has tipically been used for stereo estimation and includes methods such as random forest classifiers \cite{rf_matching}, Markov random fields (MRF) \cite{mrf_matching} and, more classically, slanted plane models \cite{slanted_matching}. In our research, we focus on a deep learning approach to the matching problem by exploiting   convolutional siamese networks \cite{matching, stereomatching}. Previous methods, however,  focused on matching pairs of small image patches.  In \cite{damnpaper} deep learning is exploited for tracking. However, this approach is only similar to our method at a very high level: using deep learning in a tracking by detection framework. Our appearance matching is based on a fully convolutional network with no requirements for optical flow and learning is done strictly via backpropagation. Furthermore, we reason in 3D and the spatial branch of our matching networks corrects for things such as ego-motion and car resemblance. In contrast \cite{damnpaper} uses optical flow and is piecewise trained using Gradient Boosting.

Tracking methods usually employ hand-crafted feature extractors with distance functions such as Chi-Square or Bhattacharyya to tackle the matching problem \cite{mincostflow,onlinecrf,oxfordt,mttonline}. In contrast, we propose to learn both the feature representations as well as the similarity with a siamese network. Furthermore, our network  takes advantage of both appearance and 3D spatial cues during matching. This is possible since we employ a 3D object detector which gives us 3D bounding boxes.

Motion models have been widely used especially in filtering based methods. \cite{motion_pf} uses a Markov random field to model motion interactions and \cite{motion_opt_flow} uses the distance between histograms of oriented optical flow (HOOF). For the scope of tracking vehicles, we have the advantage of not having to deal with severe deformations (motion-wise, vehicles can be seen as a single rigid body) or highly unpredictable behaviors (cars often simply maintain its lane, keep going forward, make controlled curves, etc), which suggests that spatial cues should be useful.

Sensory fusion approaches have been widely used in computer vision. LIDAR and camera are popular sensor sets employed in detection and tracking \cite{sensorfusion1, mv3d, sensorfusion3}. Other papers also exploit radar  \cite{sensorfusion2}.

In concurrent work \cite{deepflow} also proposes an end-to-end learned method for tracking by detection. Ours, however, exploits a structured hinge loss to backpropagate through a linear program, which simplifies the problem and yields better experimental results.
\raquel{I'm concernced about the last sentence. Did we try and compare?}
\section{Deep Structured  Tracking}
\label{sec:DSM}

In this work, we propose a novel approach to tracking by detection, which exploits the power of structure prediction as well as deep neural networks. Towards this goal, we formulate the problem as inference in a deep structured model (DSM), where the factors are computed using a set of feedforward neural nets that exploit both camera and LIDAR data to compute both detection and matching scores. Inference in the model can be done exactly by a set of feedforward processes followed by solving a linear program. Learning is done end-to-end via minimization of a structured hinge loss, optimizing simultaneously the detector and tracker. 
As shown in our experiments, this is very beneficial compared to piece-wise training.

\subsection{Model Formulation}
\label{sc:structured_model}

Given a set of candidate detections $\bx = [\obsvar_1, \obsvar_2, ..., \obsvar_k]$ estimated over a sequence of frames of arbitrary length, our goal is to estimate which detections are true positive as well as link them over time to form trajectories.
Note that this is a difficult problem since  the number of targets is unknown and can vary over time (e.g., objects can appear any time and disappear when they are no longer visible).

We parameterize the problem with four types of binary variables. For each candidate detection $x_j$ a binary variable $\binvar^{det}_{j}$  encodes if the detection is a true positive. Further, let $\binvar^{link}_{j,k}$ be a binary variable representing  if the $j$-th and $k$-th detections belong to the same object. Finally, for each detection $x_j$ two additional binary variables $\binvar^{new}_{j}$ and $\binvar^{end}_{j}$  encode whether it is the beginning or the end of a trajectory, respectively. This is necessary in order to represent the fact that certain detections are more likely to result in end of trajectory,  for example if they are close to the end of LIDAR range or if they are heavily occluded.
For notational convenience we collapse all  variables  into a vector  $\allowbreak \by = (\binvarset^{det}, \binvarset^{link}, \binvarset^{new}, \binvarset^{end})$, which comprises all candidate detections, matches, entries and exits.

We then formulate the multi-target tracking problem as an integer linear program
\begin{equation*}
\label{eq:blank_lp1}
  \begin{aligned}
  & \underset{\binvarset}{\text{maximize}}
    & & \modelfun_\weightset(\obsvarset) \binvarset \\
  & \text{subject to}
    & & A \binvarset = 0,
  \quad \by \in \{0, 1\}^{|\by|}\ 
  \end{aligned}
\end{equation*}
where $\modelfun_{\weightset}(\bx)$ is a vector comprising the cost of each random variable assignment, and $A\by=0$ is a set of constraints encoding valid trajectories, as not all assignments are possible.
We now describe the constraints and the cost function in more details.


%
%

\subsection{Conservation of Flow Constraints}
We employ a set of linear constraints (two per detection) encoding  conservation of flow in order to generate non-overlapping trajectories. This includes the fact that a detection cannot be linked to a detection belonging to the same frame. Furthermore, in order for a detection to be active, it has to either be linked to another detection in the previous frame or the trajectory should start at that point. Additionally, a detection can only end if the detection is active and not linked to another detection in the next frame. Thus, for each detection, a constraint is defined in the form of
\begin{gather}
	\label{eq:constraints}
	\binvar^{new}_j + \sum_{k \in \mathcal{N^-}(j)} \binvar^{link}_{j,k} = \binvar^{end}_{j} + \sum_{k \in \mathcal{N^+}(j)} \binvar^{link}_{j,k} = \binvar^{det}_{j} \quad \forall j
\end{gather}
where $\mathcal{N^-}(j)$ denotes the candidate detections that could be matches for the $j$-th  detection $\obsvar_j$ in the immediately preceding frame, and  $\mathcal{N^+}(j)$ in the immediately following frame.
Note that one can collapse all these constraints in matrix form to yield  $A\by = 0$.

\subsection{Deep Scoring and Matching}
\label{sec:deep_scoring}

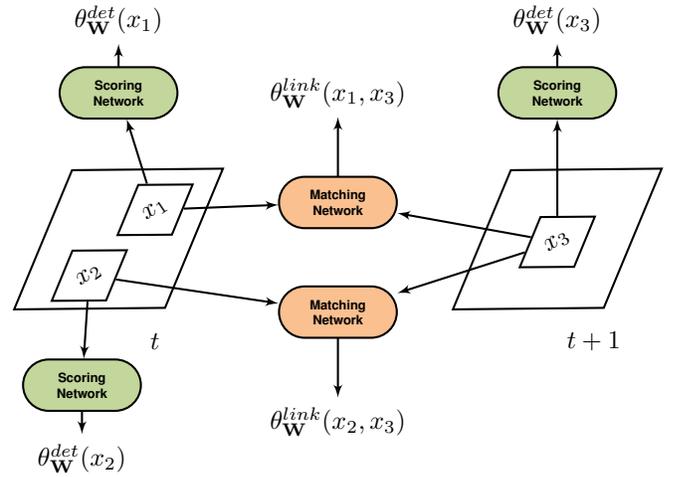
\begin{figure}[t]
  \centering
  \resizebox{\columnwidth}{!}{%
    \begin{tikzpicture}[-latex',font=\sffamily\bfseries,thick, node distance = 2cm]
    \node (F1) at (0,0) [draw, diamond, aspect=1.5, rotate=34, scale=6] {};
    \node (F1L) at (0.5,-1.4) {$t$};
    \node (F2) at (6,0) [draw, diamond, aspect=1.5, rotate=34, scale=6] {};
    \node (F1L) at (6.5,-1.4) {$t+1$};
    \node (D1) at (0.5,0.4) [draw, diamond, aspect=1.5, rotate=34, scale=1] {$x_1$};
    \node (D2) at (-0.4,-0.5) [draw, diamond, aspect=1.5, rotate=34, scale=1] {$x_2$};
    \node (D3) at (6,-0.04) [draw, diamond, aspect=1.5, rotate=34, scale=1] {$x_3$};
    \node (det1) at (0,3.) {$\modelfun_\weightset^{det}(\obsvar_1)$};
    \node (det2) at (-0.5,-3) {$\modelfun_\weightset^{det}(\obsvar_2)$};
    \node (det3) at (6,3.) {$\modelfun_\weightset^{det}(\obsvar_3)$};
    \node (M1T) at (3,2) {$\modelfun_\weightset^{link}(\obsvar_1, \obsvar_3)$};
    \node (M2T) at (3,-2.5) {$\modelfun_\weightset^{link}(\obsvar_2, \obsvar_3)$};
    \node (M1) [conv, scale=0.5, fill=matchnet] at (3,0.5) {Matching \\ Network};
    \node (M2) [conv, scale=0.5, fill=matchnet] at (3,-1) {Matching \\ Network};
    \node (D1N) [conv, scale=0.5, fill=scorenet] at (0,2) {Scoring \\ Network};
    \node (D2N) [conv, scale=0.5, fill=scorenet] at (-0.5,-2) {Scoring \\ Network};
    \node (D3N) [conv, scale=0.5, fill=scorenet] at (6,2) {Scoring \\ Network};
    \path [line] (D1) -- (M1);
    \path [line] (D3) -- (M1);
    \path [line] (M1) -- (M1T);
    \path [line] (D2) -- (M2);
    \path [line] (D3) -- (M2);
    \path [line] (M2) -- (M2T);
    \path [line] (D1) -- (D1N);
    \path [line] (D1N) -- (det1);
    \path [line] (D2) -- (D2N);
    \path [line] (D2N) -- (det2);
    \path [line] (D3) -- (D3N);
    \path [line] (D3N) -- (det3);
    \end{tikzpicture}}
  \caption{Illustration of the forward passes over a set of detections from two frames.}
  \label{fig:forward_passes}
\end{figure}

We refer the reader to \autoref{fig:forward_passes} for an illustration of the neural networks we designed for both scoring and matching. For each detection $\obsvar_j$, a forward pass of a \textbf{Detection Network} is computed to produce $\modelfun_\weightset^{det}(\obsvar_j)$, the cost of using or discarding $\obsvar_j$ according to the assignment to $y^{det}_j$. For each pair of detections $\obsvar_j$ and $\obsvar_i$ from subsequent frames, a forward pass of the \textbf{Match Network} is computed to produce $\modelfun_\weightset^{link}(\obsvar_i, \obsvar_j)$, the cost of linking or not these two detections according to the assignment to $y^{link}_{i,j}$. Finally, each detection might start a new trajectory or end an existing one, the costs for this are computed via $\modelfun_\weightset^{new}(\obsvarset)$ and $\modelfun_\weightset^{end}(\obsvarset)$, respectively, and are associated with the assignments to $y^{new}$ and $y^{end}$.
We now discuss in more details the neural networks we employed.

\subsubsection{\textbf{Detection $\modelfun_\weightset^{det}(\obsvarset)$}}
We exploit object proposals in order to reduce the search space over all possible detections.
In particular, we employ the MV3D detector \cite{mv3d} to produce oriented 3D object proposals from LIDAR and RGB data (i.e., regions in 3D where there is a high probability that a vehicle is present).
To make sure that the tracker produces accurate trajectories, we need a classifier that decides whether or not an object proposal is a true positive (i.e., actually represents a vehicle). To that end, we employ a convolutional neural network based on VGG16 \cite{vgg16} to predict whether or not there is a vehicle in the detection bounding box.
 Towards this goal, the 3D bounding boxes from MV3D are  projected onto the camera and the resulting patches are fed to the aforementioned convolutional neural network to produce detection scores.



\subsubsection{\textbf{Link $\modelfun_\weightset^{link}(\obsvarset)$}}
\label{sec:matching_network}

One of the fundamental tasks in tracking is deciding whether or not two detections in
subsequent frames represent the same object. In this work, we use deep neural networks that exploit both appearance and spatial information to represent how to match. Towards this goal, we design an architecture where one branch processes the appearance of each detection with a convolutional neural network, while two others consist of feedforward networks dealing with spatial information in 3D and 2D respectively. The activations of all branches are then fed to a fully connected layer to produce the matching score.

To extract appearance features, we employ a siamese  network based on VGG16 \cite{vgg16}. Note that in a siamese setup, the two branches (each processing a detection)   share the same set of weights. This makes the architecture more efficient  in terms of memory and allows learning with fewer examples. In particular, we resized each detection to be of dimension $224\times224$. To produce a concise representation of the activations without using fully connected layers, each of the max-pool outputs is passed through a product layer followed by a weighted sum, which produces a single scalar for each max-pool layer, yielding an activation vector of size 5. We use skip-pooling as matching should exploit both low-level features (e.g., color) as well as semantically richer features from higher layers.

To incorporate spatial information into the model, we employ fully connected architectures that model both 2D and 3D motion. In particular, we exploit 3D information in the form of  a $180\times200$ occupancy grid in bird's eye view and 2D information from the occupancy region in the frontal view camera, scaled down from the original resolution of $1242\times375$ to $124\times37$. In bird's eye perspective, each 3D detection is projected onto a ground plane, leaving only a rotated rectangle that reflects its occupancy in the world. Note that since the observer is a mobile platform (an autonomous vehicle, in this case), the coordinate system between two subsequent frames would be  shifted because the observer moved in the time elapsed. Since its speed in each axis is known from the IMU data, one can calculate the displacement of the observer between each observation and translate the coordinates accordingly. This way, both grids are on the exact same coordinate system \raquel{world coordinate system that is static?}. This approach is important to make the system invariant to the speed of the ego-car. The frontal view perspective encodes the rectangular area in the camera occupied by the target. It is the equivalent of projecting the 3D bounding box onto camera coordinates.

We use  fully connected layers to capture the spatial patterns, since vehicles behave in different ways depending on where they are with respect to the ego-car. For instance, a car in front of the observer (in the same lane) is likely to move forward, while cars on the left lane are likely to come towards the ego-car. This information would be lost in a convolutional architecture since it would be spatially invariant.

\subsubsection{\textbf{New $\modelfun_\weightset^{new}(\obsvarset)$ / End  $\modelfun_\weightset^{end}(\obsvarset)$}}

These costs are simply learned scalars intended to shift the optimization towards producing longer trajectories.

\LinesNumbered
  \SetKwInOut{Input}{Input}
\begin{algorithm}[t]
  \caption{Inference in the DSM for Tracking\label{alg:dsm_infer}}
  \Input{Input RGB+Lidar frames ($\mathbf{x}$); \newline Learned weights $\mathbf{w}$\;}
  \ForEach{Temporal window $(a,z) \in \abs{\mathbf{x}}$}{
    detections $\leftarrow$ Detector($x[a:z], \mathbf{w_{det}}$)\;
    \ForEach{Pair of linkable detections $x_i, x_j \in$ detections}
    {
      link\_score[i,j] $\leftarrow$ MatchingNet($x_i, x_j, \mathbf{w_{link}}$)\;
    }
    LP $\leftarrow$ BuildLP(detections, link\_score, $\mathbf{w_{new}}$, $\mathbf{w_{end}}$)\;
    trajectories $\leftarrow$ Optimize(LP)\;
  }
\end{algorithm}

\subsection{Inference }
\label{sc:dsm_inference}


As described before, the multi-target tracking problem can be formulated as a constrained integer programming problem. While Integer programming is NP-Hard, the constraint matrix exhibits the total unimodularity property \cite{berclaz}, which allows the problem to be relaxed to a Linear Program while still guaranteeing optimal integer solutions. Thus, we perform inference by solving
\begin{equation}
\label{eq:blank_lp}
  \begin{aligned}
  & \underset{\binvarset}{\text{maximize}}
    & & \modelfun_\weightset(\obsvarset) \binvarset   \\
  & \text{subject to}
    & & A \binvarset = 0,
  \quad \by \in [0, 1]^{|\by|}\ 
  \end{aligned}
\end{equation}
Note that other alternative formulations exist for the linear program in the form of a min cost flow problem, which can be solved via Bellmann-Ford \cite{bf_book} and Successive Shortest Paths (SSP) \cite{ssp_book}. These methods are guaranteed to give the same solution as the linear program. In this work, we simply solve the constrained linear program using  an off the shelve solver  \cite{ortools}.

Prior to solving the linear program, the costs have to be computed. This implies computing a feedforward pass from the detection network for each detection to compute $\modelfun_\weightset^{det}(\obsvarset)$, as well as a feedforward pass for every pair of linkable detections to compute $\modelfun_\weightset^{link}(\obsvarset)$. Note that $\modelfun_\weightset^{new}(\obsvarset)$ and $\modelfun_\weightset^{end}(\obsvarset)$ require no computations since they are simply learned scalars.

Once the costs are computed, the linear program can then be solved, yielding the global optimal solution for all frames.
We refer the reader to \autoref{alg:dsm_infer} for pseudocode of our approach.

\subsection{End-to-End Learning }
\label{sc:learn_dsm}

\LinesNumbered
  \SetKwInOut{Input}{Input}
\begin{algorithm}[t]
  \caption{End-to-End Learning\label{alg:dsm_learn}}
  \Input{Input RGB+LIDAR frames ($\mathbf{x}$); \newline Ground truth trajectories $\mathbf{\hat{y}}$\;}
  $\mathbf{w} \leftarrow$ initialize()\;
  \ForEach{Temporal window $(a,z) \in \abs{\mathbf{x}}$}{
    detections $\leftarrow$ Detector($x[a:z], \mathbf{w_{det}}$)\;
    \ForEach{Pair of linkable detections $x_i, x_j \in$ detections}
    {
      link\_score[i,j] $\leftarrow$ MatchingNet($x_i, x_j, \mathbf{w_{link}}$)\;
    }
    LP $\leftarrow$ BuildLossLP(detections, link\_score, $\mathbf{\hat{y}}$, $\mathbf{w_{new}}$, $\mathbf{w_{end}}$); \textit{(\autoref{eq:hinge_loss})}\\
    $\mathbf{y} \leftarrow$ Optimize(LP)\;
    grads $\leftarrow$ ComputeGradients($\mathbf{y}$, $\mathbf{\hat{y}}$)\;
    $\mathbf{w} \leftarrow$ UpdateStep($\mathbf{w}$, grads)\;
  }
\end{algorithm}

One of the main contribution of this work is an algorithm that allows us to train tracking by detection end-to-end. This is far from trivial, as it implies backpropagating through a linear program.
We capitalize on the fact that inference can be done exactly and utilize a  structured hinge loss as our loss function
\begin{equation}
  \label{eq:hinge_loss}
  \mathcal{L}(\obsvarset, \binvarset, \weightset) = \sum_{\obsvarset \in \trainset} \Bigg[ \max_\binvarset \Big(\Delta(\binvarset,\hat{\binvarset}) + \modelfun_\weightset(\obsvarset) (\binvarset - \hat{\binvarset} \Big)\Bigg]
\end{equation}
with  $\Delta(\binvarset,\hat{\binvarset})$ being the task loss representing the fact that not all mistakes are equally bad. In particular, we use the Hamming distance between the inferred variable values ($\binvarset$) and the ground truth assignments ($\hat{\binvarset}$).

We utilize subgradient descent to train our model. 
Taking the subgradients of \autoref{eq:hinge_loss} with respect to $\modelfun_\weightset(\obsvarset)$ yields
\begin{equation}
\dfrac{\partial \mathcal{L}(\obsvarset, \binvarset, \weightset)}{\partial \modelfun_\weightset(\obsvarset)} = \begin{dcases}0 & S\leq0 \\ \binvarset^* - \hat{\binvarset} & \text{otherwise}\end{dcases}
\end{equation}
where $S$ denotes the result of the summation over the batch $\trainset$ in \autoref{eq:hinge_loss}.  Furthermore, $\binvarset^*$ denotes the solution of the loss augmented inference, which in this case becomes 
\begin{equation}
\label{eq:la_lp}
  \begin{aligned}
  & \underset{\binvarset}{\text{maximize}}
    & & \modelfun_\weightset(\obsvarset) \binvarset + \Delta(\binvarset,\hat{\binvarset})\\
  & \text{subject to}
    & & A \binvarset = 0,
  \quad \by \in [0, 1]^{|\by|}\ 
  \end{aligned}
\end{equation}
As the loss decomposes this is again a LP that can be solved exactly. 

We refer the reader to \autoref{alg:dsm_learn} for a pseudocode of our end-to-end training procedure.

\section{Experimental Evaluation}
\label{sec:experimental_evaluation}

\begin{table*}[htb]
 \centering
 \begin{tabular}{llllllll}
   \hline
   Method & MOTA & MOTP & MT & ML & IDS & FRAG & FP \\\hline
   End to end & \textbf{70.66\%} & \textbf{83.08\%} & 72.17\% & 4.85\% & \textbf{54} & \textbf{239} & \textbf{1579}\\
   Piecewise & 69.02\% & 82.90\% & \textbf{74.75\%} & \textbf{3.20\%} & 97 & 289 & 1836\\
   \hline
 \end{tabular}
\caption{Comparison of tracking results between end to end and piecewise learning approaches.}
\label{tab:tracking-results}
\end{table*}

\begin{table*}[htb]
 \centering
 \begin{tabular}{lllllll}
  \hline
  Method & MOTA & MOTP & MT & ML & IDS & FRAG \\\hline
  CEM \cite{Milan2014PAMI} & 51.94 \% & 77.11 \% & 20.00 \% & 31.54 \% & 125 & 396 \\
  RMOT \cite{Yoon2015WACV} & 52.42 \% & 75.18 \% & 21.69 \% & 31.85 \% & 50 & 376 \\
  TBD \cite{Geiger2014PAMI} & 55.07 \% & 78.35 \% & 20.46 \% & 32.62 \% & 31 & 529 \\
  mbodSSP \cite{onlinemcf} & 56.03 \% & 77.52 \% & 23.23 \% & 27.23 \% & \textbf{0} & 699 \\
  SCEA \cite{Yoon2016CVPR} & 57.03 \% & 78.84 \% & 26.92 \% & 26.62 \% & 17 & 461 \\
  SSP \cite{onlinemcf} & 57.85 \% & 77.64 \% & 29.38 \% & 24.31 \% & 7 & 704 \\
  ODAMOT \cite{Gaidon2015BMVC} & 59.23 \% & 75.45 \% & 27.08 \% & 15.54 \% & 389 & 1274 \\
  NOMT-HM \cite{Choi2015ICCV} & 61.17 \% & 78.65 \% & 33.85 \% & 28.00 \% & 28 & 241 \\
  LP-SSVM \cite{Wang2016IJCV} & 61.77 \% & 76.93 \% & 35.54 \% & 21.69 \% & 16 & 422 \\
  RMOT* \cite{Yoon2015WACV} & 65.83 \% & 75.42 \% & 40.15 \% & 9.69 \% & 209 & 727 \\
  NOMT \cite{Choi2015ICCV} & 66.60 \% & 78.17 \% & 41.08 \% & 25.23 \% & 13 & \textbf{150} \\
  DCO-X* \cite{Milan2013CVPR} & 68.11 \% & 78.85 \% & 37.54 \% & 14.15 \% & 318 & 959 \\
  mbodSSP* \cite{onlinemcf} & 72.69 \% & 78.75 \% & 48.77 \% & 8.77 \% & 114 & 858 \\
  SSP* \cite{onlinemcf} & 72.72 \% & 78.55 \% & 53.85 \% & \textbf{8.00 \%} & 185 & 932 \\
  NOMT-HM* \cite{Choi2015ICCV} & 75.20 \% & 80.02 \% & 50.00 \% & 13.54 \% & 105 & 351 \\
  SCEA* \cite{Yoon2016CVPR} & 75.58 \% & 79.39 \% & 53.08 \% & 11.54 \% & 104 & 448 \\
  MDP \cite{Xiang2015ICCV} & 76.59 \% & 82.10 \% & 52.15 \% & 13.38 \% & 130 & 387 \\
  LP-SSVM* \cite{Wang2016IJCV} & 77.63 \% & 77.80 \% & 56.31 \% & 8.46 \% & 62 & 539 \\
  NOMT* \cite{Choi2015ICCV} & 78.15 \% & 79.46 \% & 57.23 \% & 13.23 \% & 31 & 207 \\
  MCMOT-CPD \cite{Lee2016ECCVWORK} & \textbf{78.90} \% & 82.13 \% & 52.31 \% & 11.69 \% & 228 & 536 \\ \hline
  DSM (ours) & 76.15 \% & \textbf{83.42 \%} & \textbf{60.00 \%} & 8.31 \% & 296 & 868 \\
  \end{tabular}
\caption{KITTI test set results.}
\label{tab:test-results}
\end{table*}

In this section, we present the performance and training details of our model. We maintain the same train/validation split as MV3D \cite{mv3d} for consistent validation results since we use this method as our detector.


\subsection{Dataset}
We use the challenging KITTI Benchmark \cite{kitti} for evaluation. This dataset consists of 40 sequences (20 for training/validation, 20 for test) with vehicles annotated in 3D. For the training set, there is a total of 8026 images and 30601 vehicles with various degrees of truncation and occlusion, the effects of which are also discussed in this section.

Since each annotated 3D trajectory contains an unique ID, it is possible to infer where trajectories begin, end and how detections are linked to form them. This allows us to determine the ground truth assignments of the binary random variables.


\begin{figure*}[!htb]
\centering
  \begin{subfigure}[b]{0.35\textwidth}
    \includegraphics[width=\columnwidth]{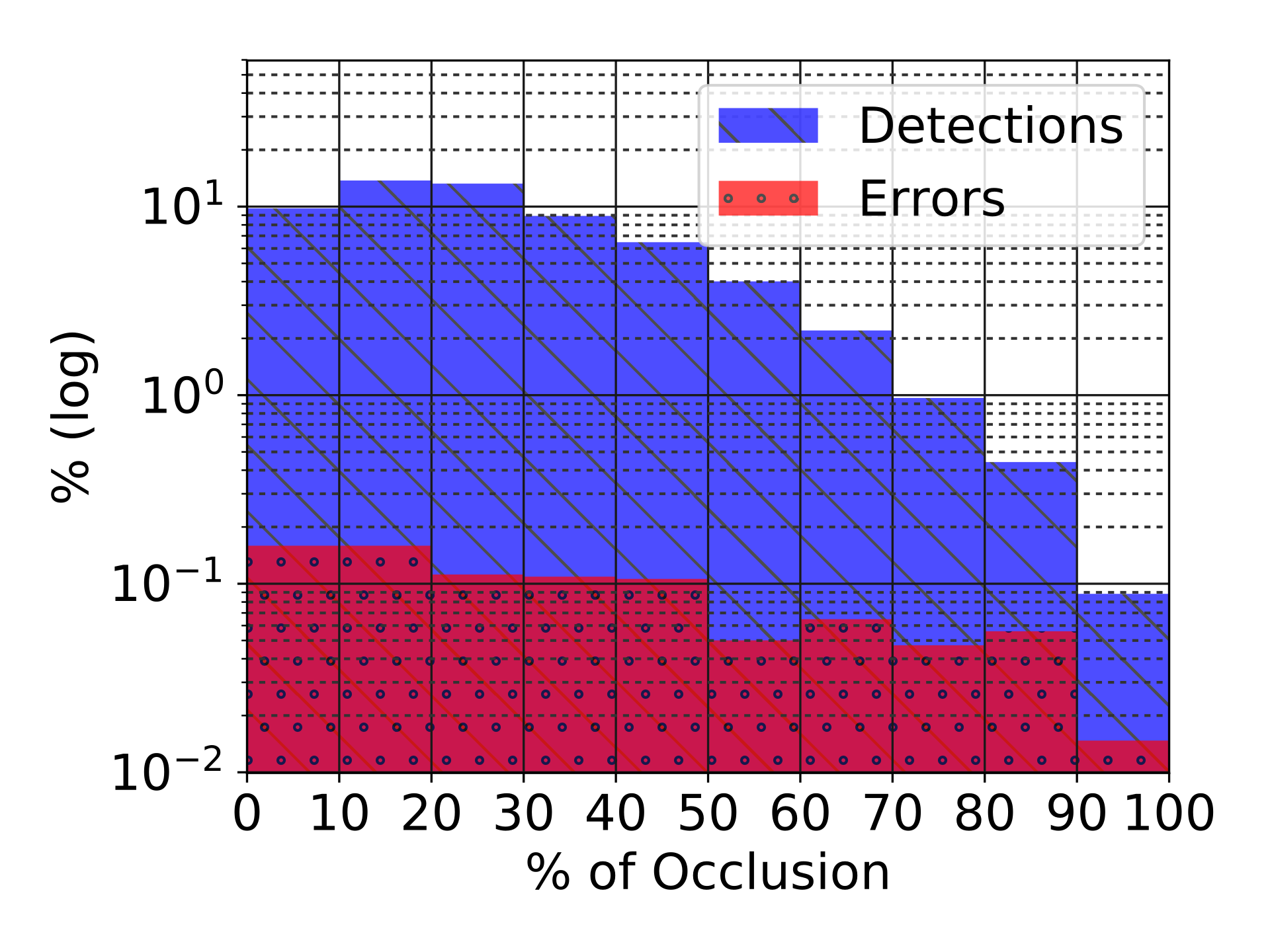}
    \caption{Occlusion.}
    \label{fig:occlusion_plot}
  \end{subfigure}
  \begin{subfigure}[b]{0.35\textwidth}
    \includegraphics[width=\columnwidth]{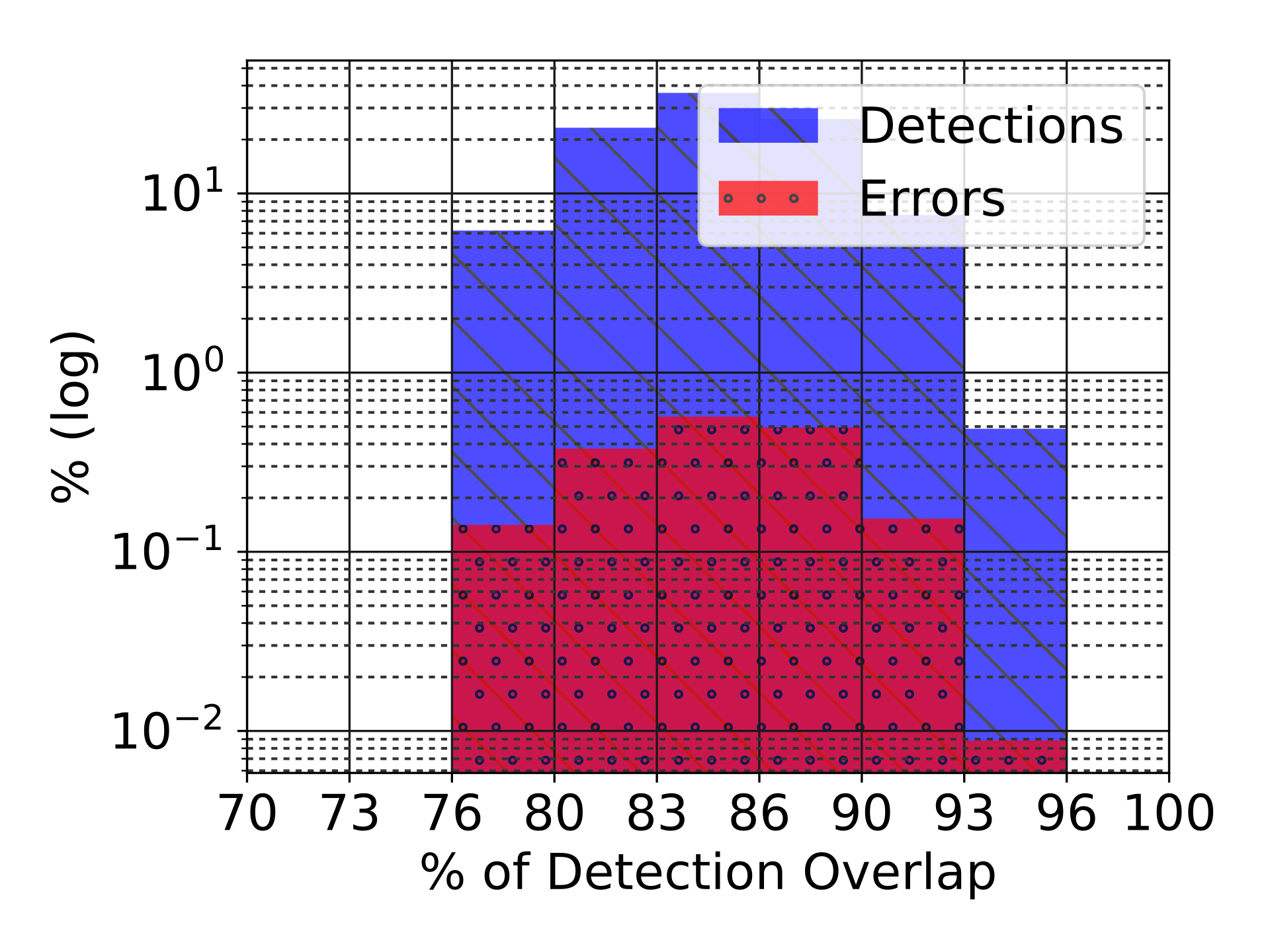}
    \caption{Precision.}
    \label{fig:precision_plot}
  \end{subfigure}
  \begin{subfigure}[b]{0.35\textwidth}
    \includegraphics[width=\columnwidth]{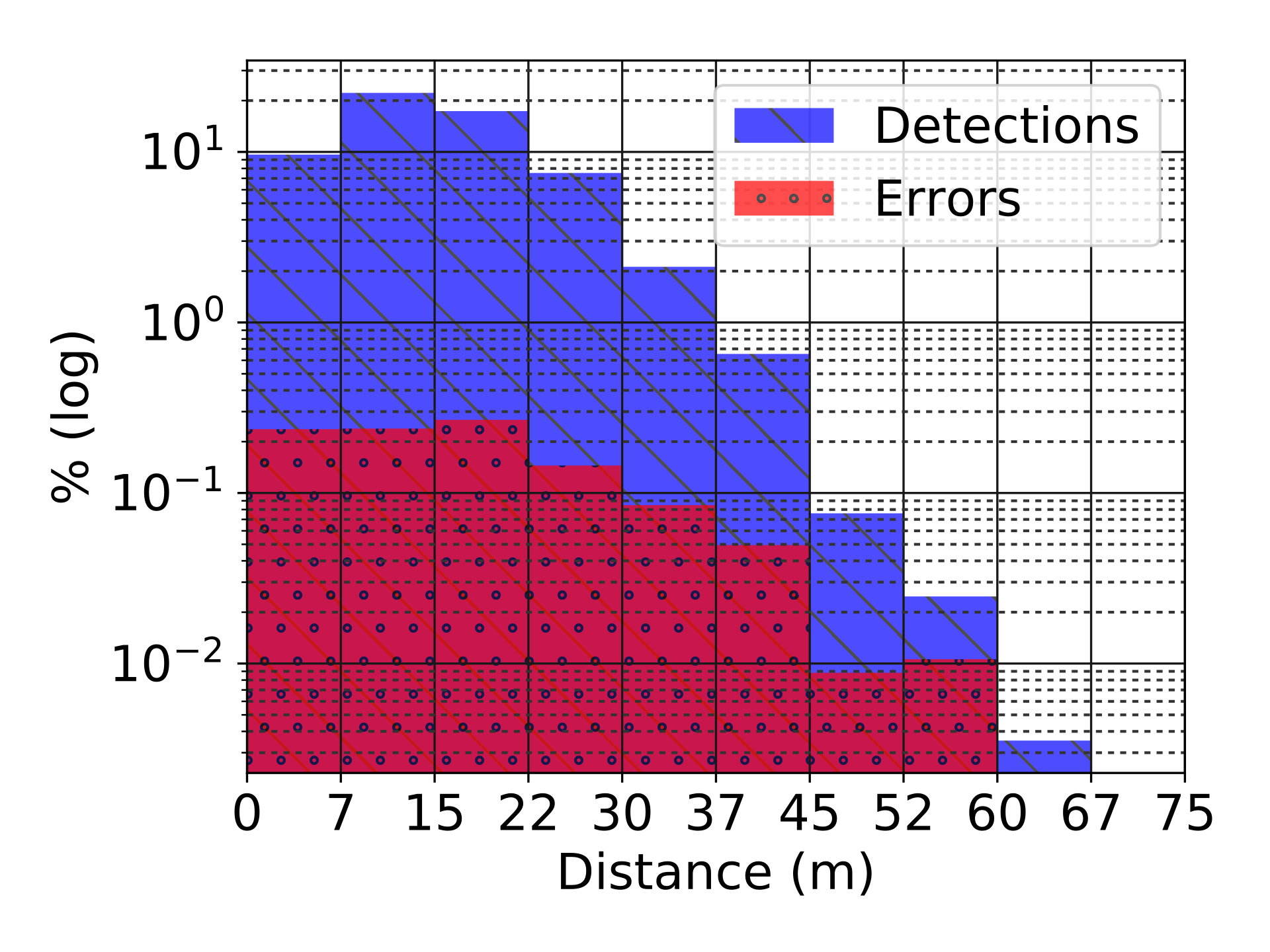}
    \caption{Distance.}
    \label{fig:distance_plot}
  \end{subfigure}
  \begin{subfigure}[b]{0.35\textwidth}
    \includegraphics[width=\columnwidth]{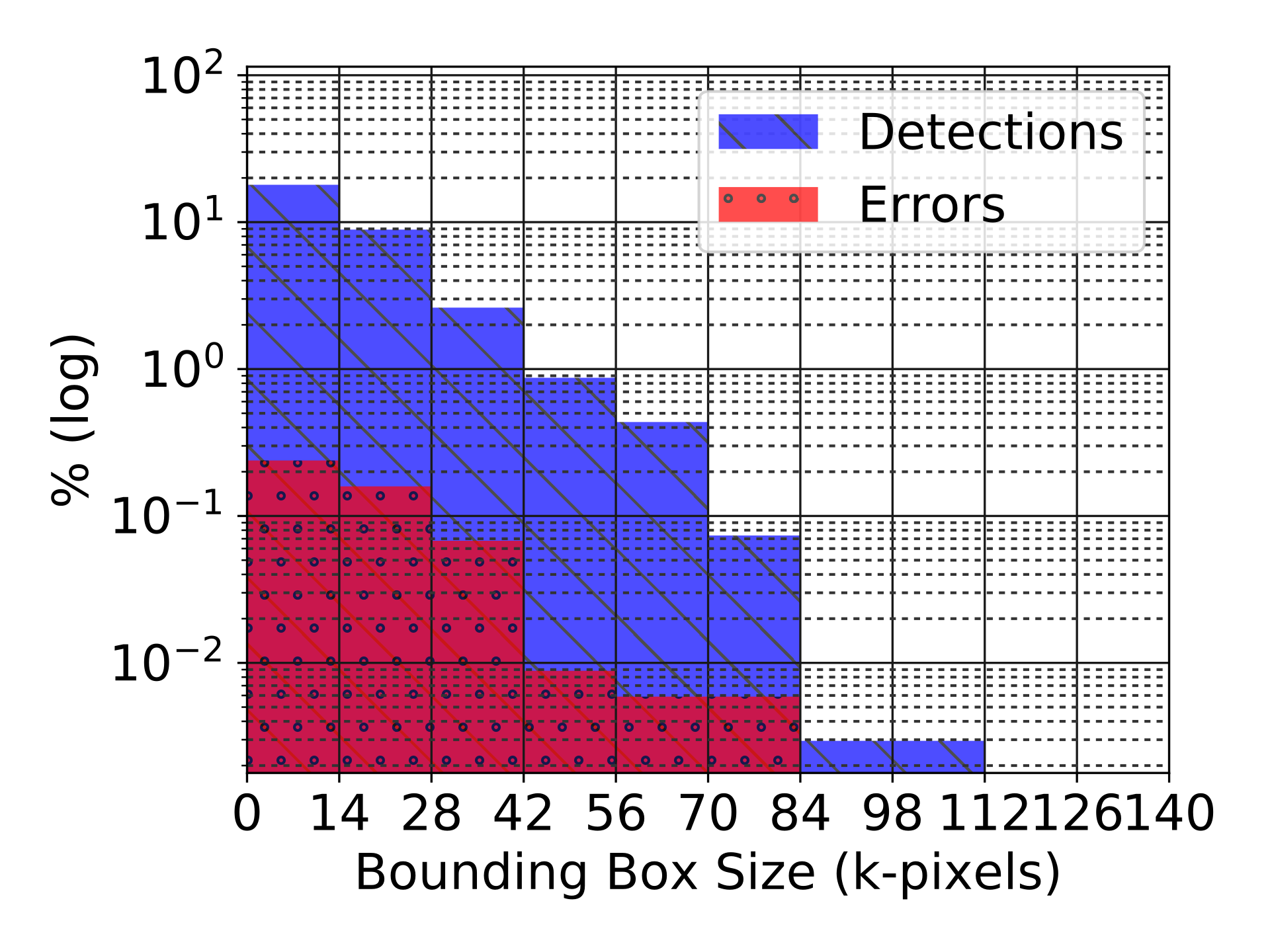}
    \caption{Bounding box size.}
    \label{fig:size_plot}
  \end{subfigure}
  \caption{Plot of detections and relative error histograms with respect to appearance conditions.}
  \label{fig:error_histograms}
\end{figure*}

\begin{figure*}[!htb]
\centering
  \begin{subfigure}[b]{.23\textwidth}
    \includegraphics[width=\columnwidth,trim={3cm 5cm 3cm 5cm},clip]{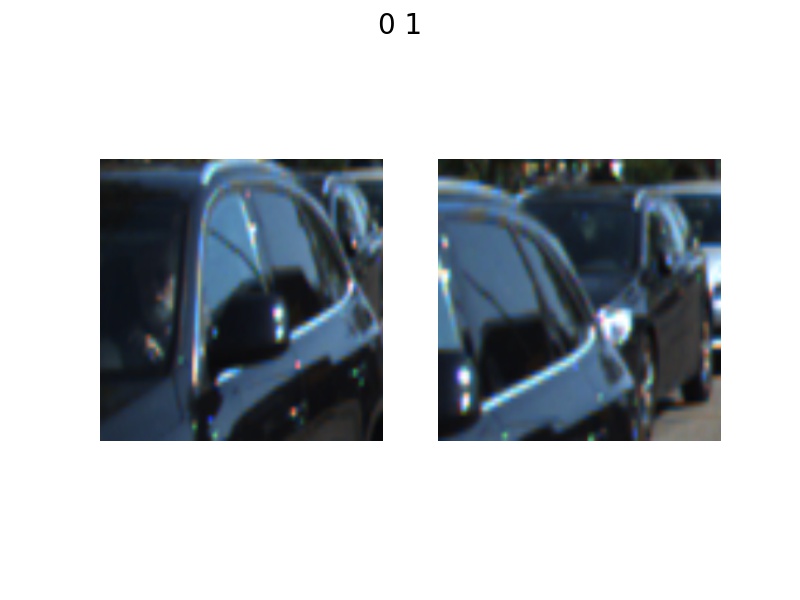}
    \caption{Occlusion}
    \label{fig:occlusion_fail}
  \end{subfigure}
  \begin{subfigure}[b]{.23\textwidth}
    \includegraphics[width=\columnwidth,trim={3cm 5cm 3cm 5cm},clip]{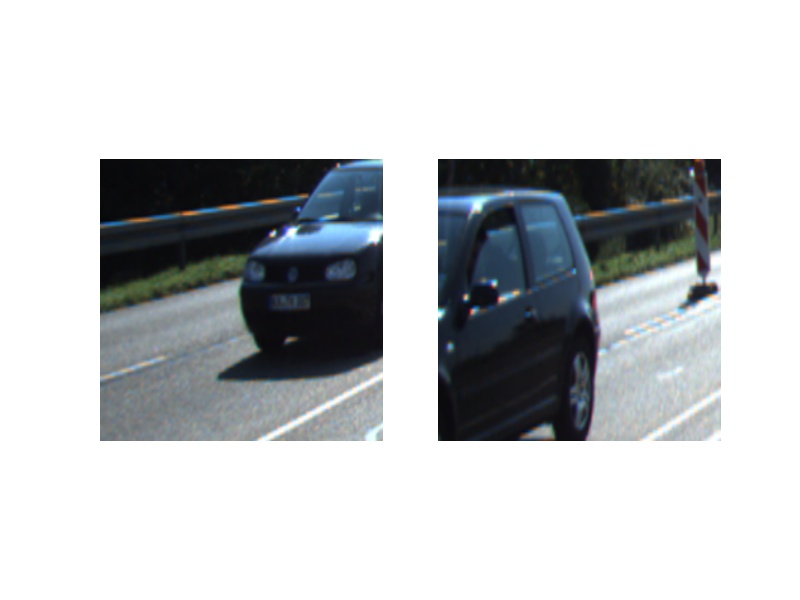}
    \caption{Truncation}
    \label{fig:truncation_fail}
  \end{subfigure}
  \begin{subfigure}[b]{.23\textwidth}
    \includegraphics[width=\columnwidth,trim={3cm 5cm 3cm 5cm},clip]{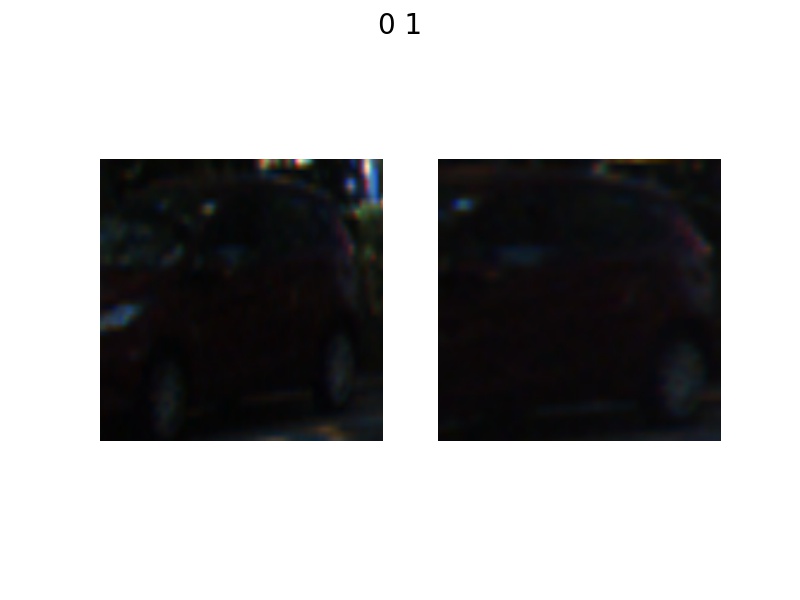}
    \caption{Lighting}
    \label{fig:visibility_fail}
  \end{subfigure}
  \begin{subfigure}[b]{.23\textwidth}
    \includegraphics[width=\columnwidth,trim={3cm 5cm 3cm 5cm},clip]{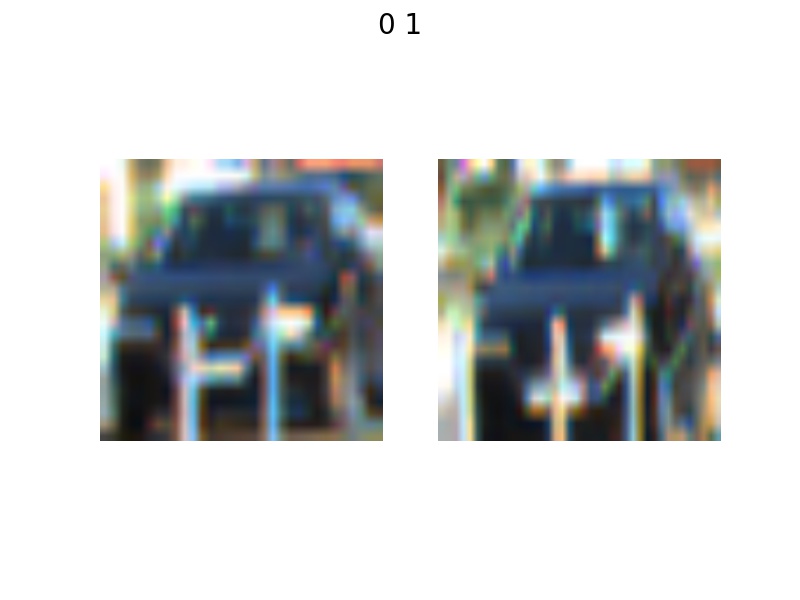}
    \caption{Distance}
    \label{fig:distance_fail}
  \end{subfigure}
  \caption{Failure modes of the matching.}
  \label{fig:matching_failures}
\end{figure*}

\subsection{Metrics}
To evaluate our matching performance we use the network accuracy when matching detections between consecutive frames. For tracking, we use the common MT/ML \cite{mttracker} metrics and CLEAR MOT \cite{clearmot} (from which we also derive ID-Switches, Fragmentations and False Positives). We refer the reader to the references for an in-depth explanation of the metrics. For completeness, we also add a brief explanation.

The MOT metric accounts for tracker accuracy (MOTA) and precision (MOTP). Accuracy measures errors in the trajectory configuration: misses, false positives and mismatches. It gives a measure of how well the tracker is able to detect objects and keep consistent trajectories. Precision measures the total error in estimated position between object-hypotheses pairs. It evaluates the tracker's ability to estimate precise object positions.

ID-Switches (IDS) account for the number of times a trajectory switches its ground-truth ID. Meanwhile, a fragmentation (FRAG) happens when part of a trajectory is not covered (usually due to missing detections). Lastly, a false-positive (FP) is a detection that does not correspond to any ground-truth bounding box. Note that in the KITTI benchmark all these metrics are computed in 2D, which does not fully evaluate our method since no evaluation is done with respect to the 3D positioning of the trajectories.

MT/ML evaluate how well the tracker is able to follow an object. A trajectory is considered mostly tracked (MT) if more than 80\% of its ground-truth length is covered by an estimated trajectory. It is considered mostly lost (ML) when it is covered for less than 20\% of its length. These metrics account for the percentage of trajectories that fall in each category.

\subsection{Training Parameters}

We use Adam optimizer \cite{kingma2014adam} with a learning rate of $10^{-5}$, $\beta_1$ of $0.9$ and $\beta_2$ of $0.999$. The CNNs are initialized with the pre-trained VGG16 weights on ImageNet and the fully connected layers (which includes the weights of the binary random variables $\binvarset$) are initialized by sampling from a truncated normal distribution (a normal distribution in which values sampled more than 2 standard deviations from the mean are dropped and re-picked) with 0 mean and $10^{-3}$ standard deviation.

\subsection{Experiments}

\vspace{0.1cm}
\noindent{\bf{Comparison to Piecewise Training:}}
First, we evaluate the importance of training  end-to-end. To that end, we compare two instantiations of our model. The first one is trained end-to-end while the second one is trained in a piecewise manner. As shown in  \autoref{tab:tracking-results} end-to-end training  outperforms  piecewise training in the metrics that we optimize for,  i.e., precision and accuracy, while showing a decrease in coverage. This is explained by the fact that the network will perform better for the task it is trained for. Furthermore, there is an inherent trade-off between coverage and accuracy. The way our cost is defined pulls the model towards producing shorter but accurate trajectories (maximize MOTA and minimize ID-switches). We note that this is better in the context of autonomous driving, as merging different tracks on the same vehicle can produce very inaccurate velocity estimates, resulting in possible collisions.

\vspace{0.1cm}
\noindent{\bf{Comparison to State of the Art}:}
In \autoref{tab:test-results} we compare our model to publicly available methods in the KITTI Tracking Benchmark. The performance of out approach is competitive with the state of the art, outperforming all other methods in some of the metrics (best for MOTP and MT, second best for ML). Furthermore, it is worth noting that tracking performance is highly correlated with detection performance in all  tracking by detection approaches.

\begin{figure*}[!htb]
\centering
  \begin{subfigure}[b]{0.33\textwidth}
    \includegraphics[width=\columnwidth]{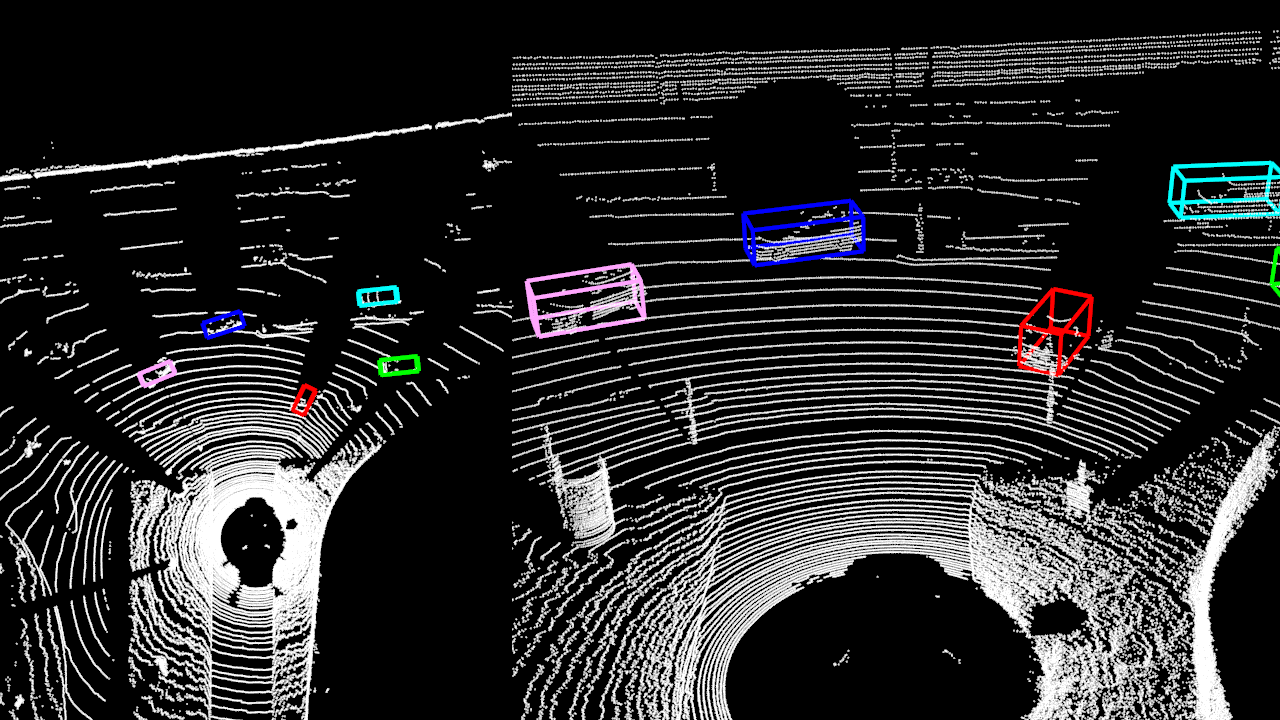}
  \end{subfigure}
  \begin{subfigure}[b]{0.56\textwidth}
    \includegraphics[width=\columnwidth]{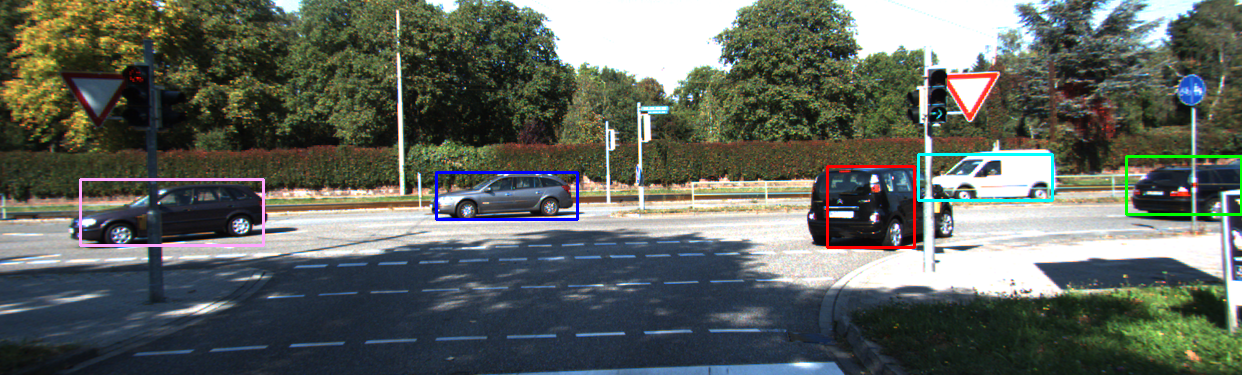}
  \end{subfigure}
  \begin{subfigure}[b]{0.33\textwidth}
    \includegraphics[width=\columnwidth]{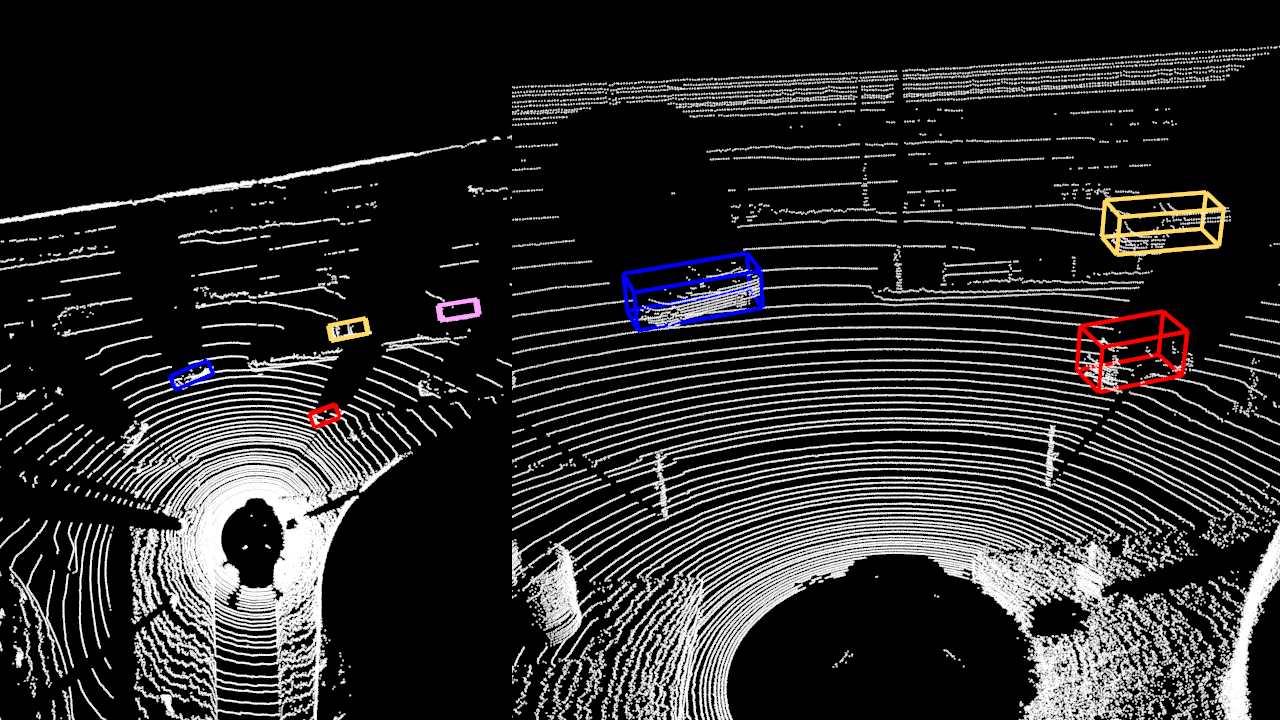}
  \end{subfigure}
  \begin{subfigure}[b]{0.56\textwidth}
    \includegraphics[width=\columnwidth]{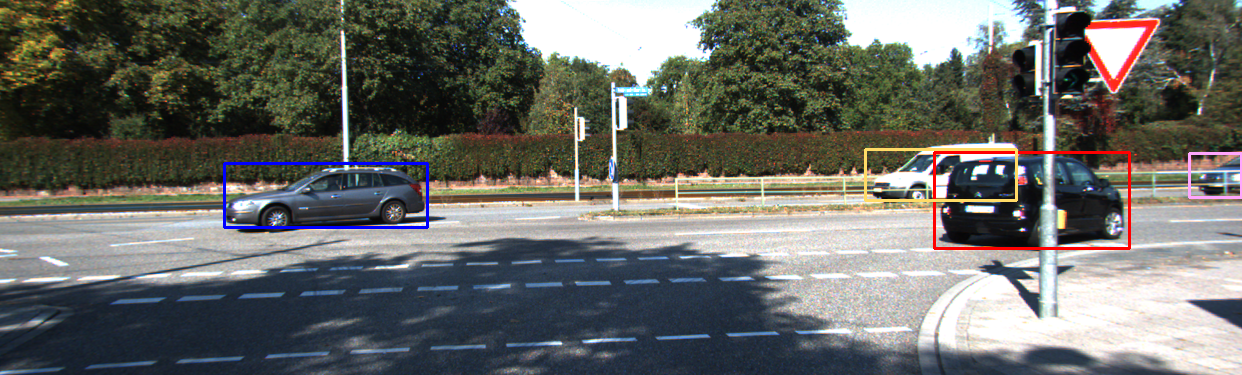}
  \end{subfigure}
  \begin{subfigure}[b]{0.33\textwidth}
    \includegraphics[width=\columnwidth]{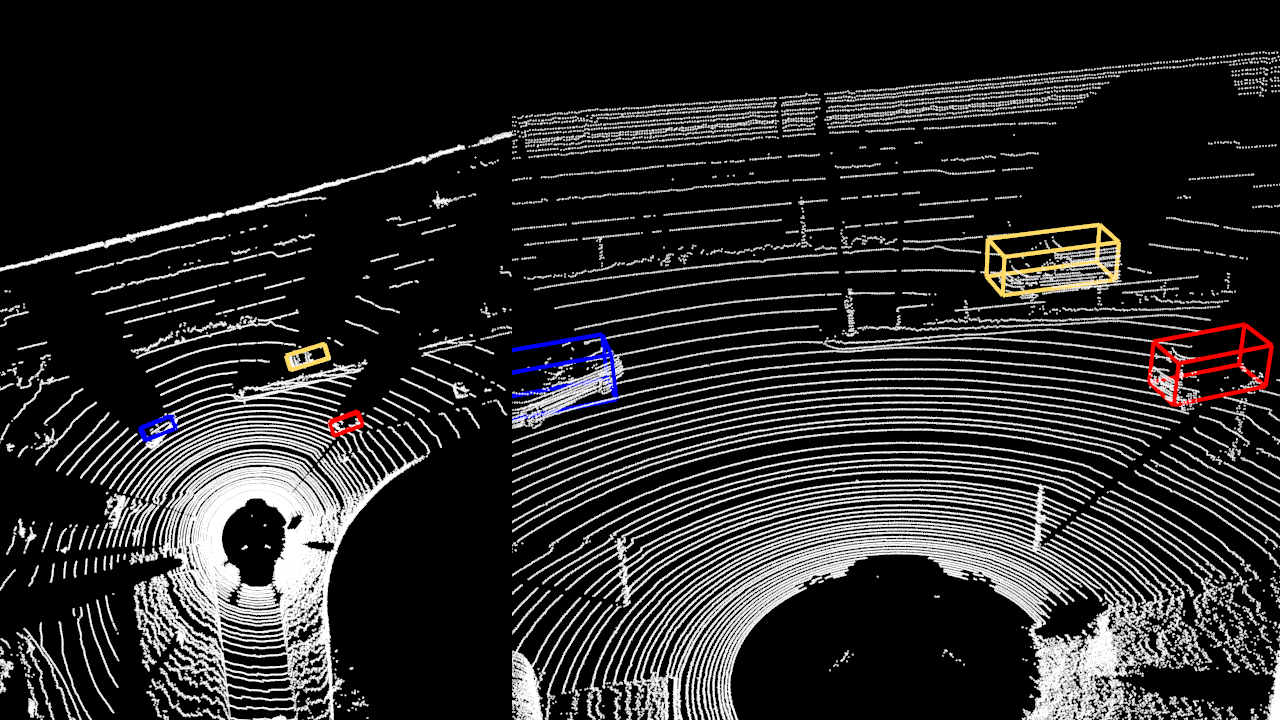}
  \end{subfigure}
  \begin{subfigure}[b]{0.56\textwidth}
    \includegraphics[width=\columnwidth]{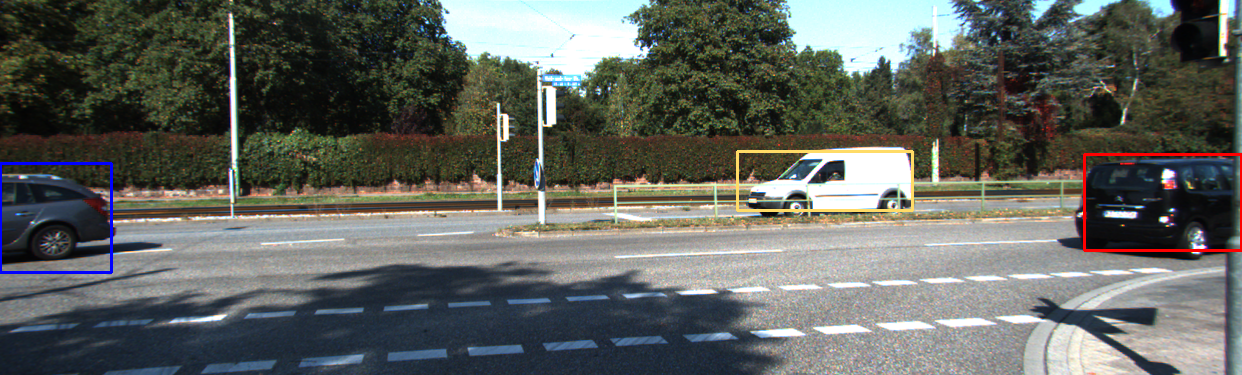}
  \end{subfigure}
  \begin{subfigure}[b]{0.33\textwidth}
    \includegraphics[width=\columnwidth]{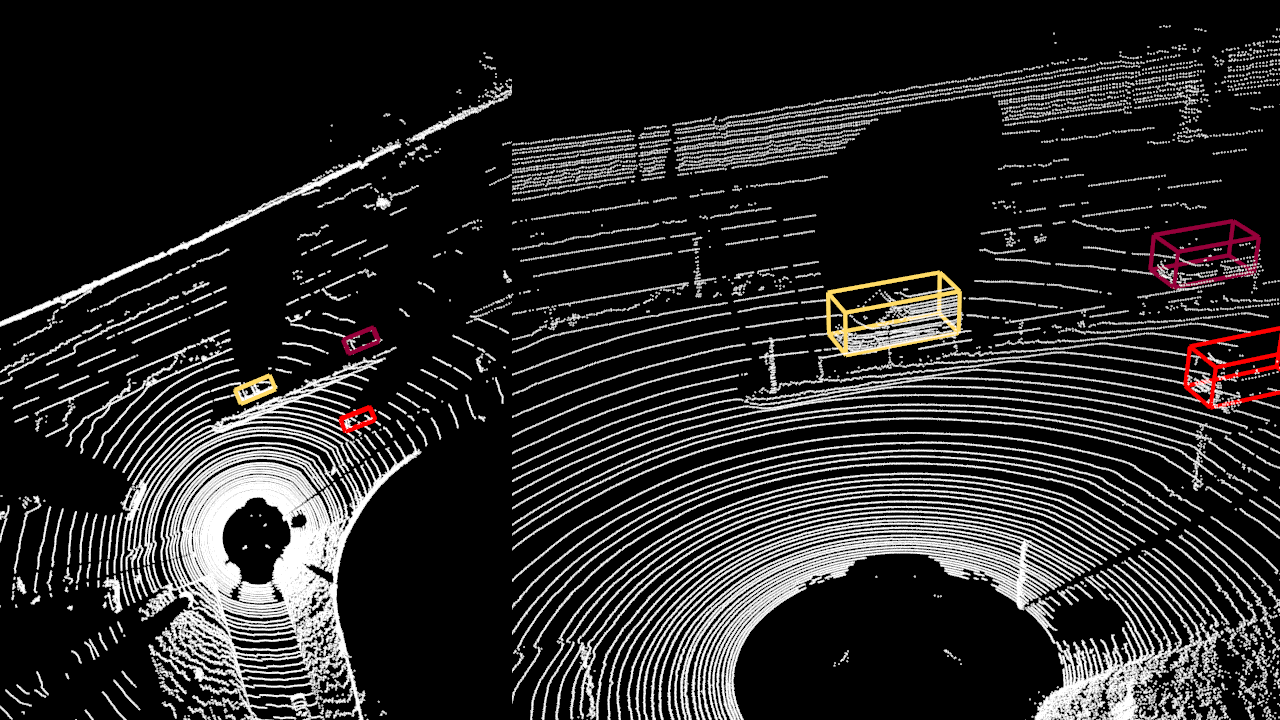}
  \end{subfigure}
  \begin{subfigure}[b]{0.56\textwidth}
    \includegraphics[width=\columnwidth]{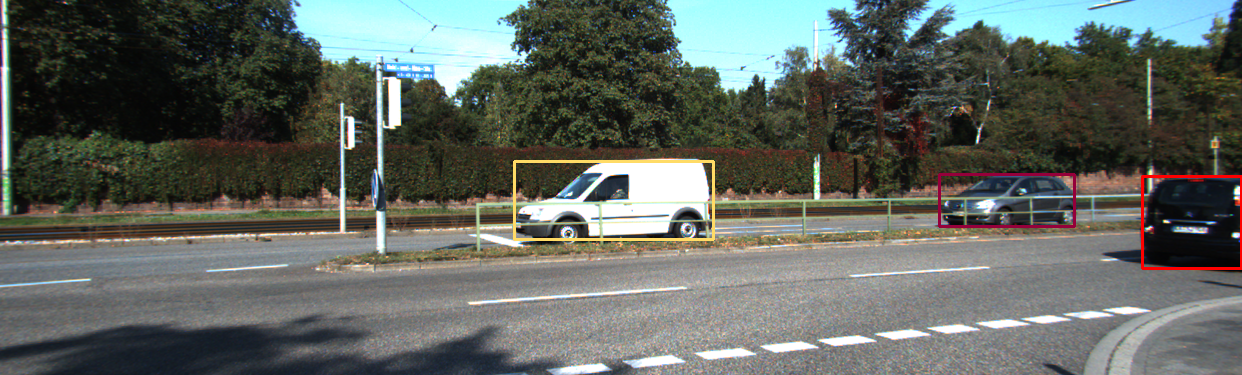}
  \end{subfigure}
  \caption{Visualization of a set of trajectories produced by the tracker over 15 frames. Trajectories are color coded, such that having the same color means it's the same object.}
  \label{fig:quali_results}
\end{figure*}

We also reiterate that our method performs tracking in 3D (which is not the case in the other  methods) and KITTI only evaluates the metrics in 2D, which does not fully represent the performance of the approach. We refer the reader to \autoref{fig:quali_results} for an example of the trajectories produced by our tracker.

\vspace{0.1cm}
\noindent{\bf{Matching Performance}:}
To validate the efficacy of our matching network, we compare it against common afinity metrics used in the literature. In particular, we evaluate methods that operate in image space by computing the distance between the distribution of colors in two detections according to the Bhattacharyya, Chi-Square and Correlation matrix. We also evaluate spatial metrics that compute the overlap, position distance, size similarity and orientation similarity between two detections.
The comparison results are shown in \autoref{tab:matching-results}. Note that the binary thresholds for the affinity metrics are defined via cross validation using for consistency the same train/validation split as the matching network. The results show that our proposed approach significantly outperforms all affinity metrics presented. \raquel{you need to explain what these baselines are!}

\vspace{0.1cm}
\noindent{\bf{Error Analysis}:}
To support our claim, we plot the error rate of the matching network as a function of a series of operating conditions in \autoref{fig:error_histograms}, all of them with the y-axis scaled logarithmically. For each bin in the histogram, we plot the percentage of detections which fall in that category and how many of them are mismatched.

\raquel{Is this error analyis wrt detector? I thougth this was the tracker...}

In \autoref{fig:occlusion_plot}, it is observable how occlusions make the task of matching increasingly harder up until 50\%, when it then settles. It is also worth noting that objects are often occluded to some extent in the dataset, considering how close to uniform the distribution is.

The effects of the detector's precision are evaluated in \autoref{fig:precision_plot}, where the performance is plotted against the detection overlap percentage, defined as the intersection over union between the detection and its ground-truth pair. Notice that the relative error rate remains close to constant in this range, which suggests that the matching network is robust to the tightness of the bounding box.

Finally, the performance with respect to bounding box dimension and object distance is analyzed. \autoref{fig:distance_plot} shows the error rate against the object distance and suggests that far away vehicles are harder to match since less information is captured by the camera. The size histogram of \autoref{fig:size_plot} corroborates this, considering that the size of the bounding box in 2D is directly correlated to its distance in 3D.

In \autoref{fig:matching_failures} failure modes of the matching network are shown, in which there are cases where matching fails due to a car being partially occluded \ref{fig:occlusion_fail}, truncated \ref{fig:truncation_fail}, poorly lit \ref{fig:visibility_fail} or too far away \ref{fig:distance_fail}.

\begin{table}[t]
 \centering
 \begin{tabular}{l|lll}
   \hline
   Method & Error\\ \hline
   Cosine Similarity & 29.66\%\\
   Color Correlation & 16.31\%\\
   Bhattacharyya & 11.93\% \\
   Chi Square & 8.02\% \\
   Bounding Box Size & 7.31\% \\
   Bounding Box Position & 6.13\% \\
   Bounding Box Overlap & 5.30\% \\\hline
   Deep Matching (ours) & \textbf{3.27\%} \\
 \end{tabular}
 \caption{Comparison to other matching methods.}
\label{tab:matching-results}
\vspace{-0.05cm}
\end{table}


\section{Conclusions}
\label{sec:conclusion}

We have proposed a novel approach to tracking by detection, which exploits the power of structure prediction as well as deep neural networks. Towards this goal, we formulate the problem as inference in a deep structured model (DSM), where the factors are computed using a set of feedforward neural nets that exploit both camera and LIDAR data to compute both detection and matching scores. Inference in the model can be done exactly by a set of feedforward processes followed by solving a linear program. Learning is done end-to-end via minimization of a structured hinge loss, optimizing simultaneously the detector and tracker.
Experimental evaluation on the challenging KITTI dataset show that our approach is very competitive outperforming the state of the art in the MOTP and MT metrics \raquel{say which metrics.} 
In the future, we plan to  extend our  approach to handle long-term occlusions and missing detections.

\bibliography{references}

\begin{thebibliography}{10}
\providecommand{\url}[1]{#1}
\csname url@samestyle\endcsname
\providecommand{\newblock}{\relax}
\providecommand{\bibinfo}[2]{#2}
\providecommand{\BIBentrySTDinterwordspacing}{\spaceskip=0pt\relax}
\providecommand{\BIBentryALTinterwordstretchfactor}{4}
\providecommand{\BIBentryALTinterwordspacing}{\spaceskip=\fontdimen2\font plus
\BIBentryALTinterwordstretchfactor\fontdimen3\font minus
  \fontdimen4\font\relax}
\providecommand{\BIBforeignlanguage}[2]{{%
\expandafter\ifx\csname l@#1\endcsname\relax
\typeout{** WARNING: IEEEtran.bst: No hyphenation pattern has been}%
\typeout{** loaded for the language `#1'. Using the pattern for}%
\typeout{** the default language instead.}%
\else
\language=\csname l@#1\endcsname
\fi
#2}}
\providecommand{\BIBdecl}{\relax}
\BIBdecl

\bibitem{onlinemcf}
P.~Lenz, A.~Geiger, and R.~Urtasun, ``{FollowMe}: Efficient online min-cost
  flow tracking with bounded memory and computation,'' \emph{IEEE Conference on
  Computer Vision and Pattern Recognition}, pp. 4364--4372, 2015.

\bibitem{thrun2005probabilistic}
S.~Thrun, W.~Burgard, and D.~Fox, ``Probabilistic robotics,'' 2005.

\bibitem{urtasun20063d}
R.~Urtasun, D.~J. Fleet, and P.~Fua, ``3d people tracking with gaussian process
  dynamical models,'' \emph{IEEE Conference on Computer Vision and Pattern
  Recognition}, vol.~1, pp. 238--245, 2006.

\bibitem{Collins2014}
R.~T. Collins and P.~Carr, ``Hybrid stochastic/deterministic optimization for
  tracking sports players and pedestrians,'' \emph{European Conference on
  Computer Vision}, pp. 298--313, 2014.

\bibitem{choi_pami13}
W.~Choi, C.~Pantofaru, and S.~Savarese, ``A general framework for tracking
  multiple people from a moving camera,'' \emph{IEEE Transactions on Pattern
  Analysis and Machine Intelligence}, vol.~35, no.~7, pp. 1577--1591, 2013.

\bibitem{berclaz}
J.~Berclaz, F.~Fleuret, E.~Turetken, and P.~Fua, ``Multiple object tracking
  using k-shortest paths optimization,'' \emph{IEEE Transactions on Pattern
  Analysis and Machine Intelligence}, vol.~33, no.~9, pp. 1806--1819, 2011.

\bibitem{shitrit}
H.~B. Shitrit, J.~Berclaz, F.~Fleuret, and P.~Fua, ``Multi-commodity network
  flow for tracking multiple people.'' \emph{IEEE Transactions on Pattern
  Analysis and Machine Intelligence}, vol.~36, no.~8, pp. 1614--1627, 2014.

\bibitem{mincostflow}
L.~Zhang, Y.~Li, and R.~Nevatia, ``Global data association for multi-object
  tracking using network flows,'' \emph{IEEE Conference on Computer Vision and
  Pattern Recognition}, pp. 1--8, 2008.

\bibitem{onlinecrf}
B.~Yang and R.~Nevatia, ``An online learned {CRF} model for multi-target
  tracking,'' \emph{IEEE Conference on Computer Vision and Pattern
  Recognition}, pp. 2034--2041, 2012.

\bibitem{milan2016online}
A.~Milan, S.~H. Rezatofighi, A.~Dick, I.~Reid, and K.~Schindler, ``Online
  multi-target tracking using recurrent neural networks,''
  \emph{arXiv:1604.03635}, 2016.

\bibitem{rf_matching}
R.~Haeusler, R.~Nair, and D.~Kondermann, ``Ensemble learning for confidence
  measures in stereo vision,'' \emph{IEEE Conference on Computer Vision and
  Pattern Recognition}, pp. 305--312, 2013.

\bibitem{mrf_matching}
A.~Spyropoulos, N.~Komodakis, and P.~Mordohai, ``Learning to detect ground
  control points for improving the accuracy of stereo matching,'' \emph{IEEE
  Conference on Computer Vision and Pattern Recognition}, pp. 1621--1628, 2014.

\bibitem{slanted_matching}
S.~Birchfield and C.~Tomasi, ``Multiway cut for stereo and motion with slanted
  surfaces.'' \emph{IEEE International Conference on Computer Vision}, pp.
  489--495, 1999.

\bibitem{matching}
J.~Zbontar and Y.~LeCun, ``Computing the stereo matching cost with a
  convolutional neural network,'' \emph{IEEE Conference on Computer Vision and
  Pattern Recognition}, pp. 1592--1599, 2015.

\bibitem{stereomatching}
W.~Luo, A.~G. Schwing, and R.~Urtasun, ``Efficient deep learning for stereo
  matching,'' \emph{IEEE Conference on Computer Vision and Pattern
  Recognition}, pp. 5695--5703, 2016.

\bibitem{damnpaper}
L.~Leal-Taix{\'e}, C.~Canton-Ferrer, and K.~Schindler, ``Learning by tracking:
  Siamese cnn for robust target association,'' \emph{IEEE Conference on
  Computer Vision and Pattern Recognition Workshops}, pp. 33--40, 2016.

\bibitem{oxfordt}
B.~Benfold and I.~Reid, ``Stable multi-target tracking in real-time
  surveillance video,'' \emph{IEEE Conference on Computer Vision and Pattern
  Recognition}, pp. 3457--3464, 2011.

\bibitem{mttonline}
C.~hao Kuo, C.~Huang, and R.~Nevatia, ``Multi-target tracking by on-line
  learned discriminative appearance models,'' \emph{IEEE Conference on Computer
  Vision and Pattern Recognition}, pp. 685--692, 2010.

\bibitem{motion_pf}
Z.~Khan, T.~Balch, and F.~Dellaert, ``Efficient particle filter-based tracking
  of multiple interacting targets using an {MRF}-based motion model,''
  \emph{IEEE Conference on Intelligent Robots and Systems}, pp. 254--259, 2003.

\bibitem{motion_opt_flow}
D.~Riahi and G.~Bilodeau, ``Multiple object tracking based on sparse generative
  appearance modeling,'' \emph{IEEE Conference on Image Processing}, pp.
  4017--4021, 2015.

\bibitem{sensorfusion1}
H.~Weigel, P.~Lindner, and G.~Wanielik, ``Vehicle tracking with lane assignment
  by camera and lidar sensor fusion,'' in \emph{IEEE Intelligent Vehicles
  Symposium}, June 2009, pp. 513--520.

\bibitem{mv3d}
X.~Chen, H.~Ma, J.~Wan, B.~Li, and T.~Xia, ``Multi-view 3d object detection
  network for autonomous driving,'' \emph{arXiv:1611.07759}, 2016.

\bibitem{sensorfusion3}
F.~Zhang, D.~Clarke, and A.~Knoll, ``Vehicle detection based on lidar and
  camera fusion,'' in \emph{IEEE Conference on Intelligent Transportation
  Systems}, Oct 2014, pp. 1620--1625.

\bibitem{sensorfusion2}
R.~O. Chavez-Garcia and O.~Aycard, ``Multiple sensor fusion and classification
  for moving object detection and tracking,'' \emph{IEEE Transactions on
  Intelligent Transportation Systems}, vol.~17, no.~2, pp. 525--534, Feb 2016.

\bibitem{deepflow}
S.~Schulter, P.~Vernaza, W.~Choi, and M.~Chandraker, ``Deep network flow for
  multi-object tracking,'' \emph{IEEE Conference on Computer Vision and Pattern
  Recognition}, pp. 2730--2739, 2017.

\bibitem{vgg16}
K.~Simonyan and A.~Zisserman, ``Very deep convolutional networks for
  large-scale image recognition,'' \emph{arXiv:1409.1556}, 2014.

\bibitem{bf_book}
D.~P. Bertsekas, R.~G. Gallager, and P.~Humblet, ``Data networks,'' vol.~2,
  1992.

\bibitem{ssp_book}
R.~K. Ahuja, T.~L. Magnanti, and J.~B. Orlin, ``Network flows: theory,
  algorithms, and applications,'' 1993.

\bibitem{ortools}
\BIBentryALTinterwordspacing
``{Google Optimization Tools},'' 2017. [Online]. Available:
  \url{https://developers.google.com/optimization/}
\BIBentrySTDinterwordspacing

\bibitem{Milan2014PAMI}
A.~Milan, S.~Roth, and K.~Schindler, ``Continuous energy minimization for
  multitarget tracking,'' \emph{IEEE Transactions on Pattern Analysis and
  Machine Intelligence}, vol.~36, no.~1, pp. 58--72, 2014.

\bibitem{Yoon2015WACV}
J.~H. Yoon, M.-H. Yang, J.~Lim, and K.-J. Yoon, ``Bayesian multi-object
  tracking using motion context from multiple objects,'' \emph{IEEE Winter
  Conference on Applications of Computer Vision}, pp. 33--40, 2015.

\bibitem{Geiger2014PAMI}
A.~Geiger, M.~Lauer, C.~Wojek, C.~Stiller, and R.~Urtasun, ``3d traffic scene
  understanding from movable platforms,'' \emph{IEEE Transactions on Pattern
  Analysis and Machine Intelligence}, vol.~36, no.~5, pp. 1012--1025, 2014.

\bibitem{Yoon2016CVPR}
J.~Hong~Yoon, C.-R. Lee, M.-H. Yang, and K.-J. Yoon, ``Online multi-object
  tracking via structural constraint event aggregation,'' \emph{IEEE Conference
  on Computer Vision and Pattern Recognition}, pp. 1392--1400, 2016.

\bibitem{Gaidon2015BMVC}
A.~Gaidon and E.~Vig, ``Online domain adaptation for multi-object tracking,''
  \emph{arXiv:1508.00776}, 2015.

\bibitem{Choi2015ICCV}
W.~Choi, ``Near-online multi-target tracking with aggregated local flow
  descriptor,'' \emph{IEEE International Conference on Computer Vision}, pp.
  3029--3037, 2015.

\bibitem{Wang2016IJCV}
S.~Wang and C.~C. Fowlkes, ``Learning optimal parameters for multi-target
  tracking.'' \emph{British Machine Vision Conference}, pp. 4--1, 2015.

\bibitem{Milan2013CVPR}
A.~Milan, K.~Schindler, and S.~Roth, ``Detection-and trajectory-level exclusion
  in multiple object tracking,'' \emph{IEEE Conference on Computer Vision and
  Pattern Recognition}, pp. 3682--3689, 2013.

\bibitem{Xiang2015ICCV}
Y.~Xiang, A.~Alahi, and S.~Savarese, ``Learning to track: Online multi-object
  tracking by decision making,'' \emph{IEEE International Conference on
  Computer Vision}, pp. 4705--4713, 2015.

\bibitem{Lee2016ECCVWORK}
B.~Lee, E.~Erdenee, S.~Jin, M.~Y. Nam, Y.~G. Jung, and P.~K. Rhee,
  ``Multi-class multi-object tracking using changing point detection,''
  \emph{European Conference on Computer Vision}, pp. 68--83, 2016.

\bibitem{kitti}
A.~Geiger, P.~Lenz, and R.~Urtasun, ``Are we ready for autonomous driving? the
  {KITTI} vision benchmark suite,'' \emph{IEEE Conference on Computer Vision
  and Pattern Recognition}, pp. 3354--3361, 2012.

\bibitem{mttracker}
Y.~Li, C.~Huang, and R.~Nevatia, ``Learning to associate: Hybridboosted
  multi-target tracker for crowded scene,'' \emph{IEEE Conference on Computer
  Vision and Pattern Recognition}, pp. 2953--2960, 2009.

\bibitem{clearmot}
K.~Bernardin and R.~Stiefelhagen, ``Evaluating multiple object tracking
  performance: the clear mot metrics,'' \emph{EURASIP Journal on Image and
  Video Processing}, vol. 2008, no.~1, pp. 1--10, 2008.

\bibitem{kingma2014adam}
D.~P. Kingma and J.~Ba, ``Adam: A method for stochastic optimization,''
  \emph{arXiv:1412.6980}, 2014.

\end{thebibliography}

\end{document}